\documentclass[sigconf]{acmart}

\AtBeginDocument{%
  \providecommand\BibTeX{{%
    \rm B\kern-.05em{\sc i\kern-.025em b}\kern-.08em\TeX}}}

\setcopyright{acmcopyright}
\copyrightyear{2024} 
\acmYear{2024} 
\acmDOI{XXXXXXX.XXXXXXX} 

\acmConference[Conference acronym 'XX]{Make sure to enter the correct
  conference title from your rights confirmation email}{June 03--05,
  2024}{City, State} 
\acmBooktitle{Proceedings of the ... Conference (Conference acronym 'XX),
 June 03--05, 2024, City, State} 
\acmPrice{15.00} 
\acmISBN{978-1-4503-XXXX-X/24/06} 

\graphicspath{{figures/}{pictures/}{images/}{./}} 

\usepackage{tabu}                      
\usepackage{booktabs}                  
\usepackage{enumitem}
\setlist{nosep} 
\usepackage{subcaption}
\usepackage{svg} 
\usepackage{tabularx}
\usepackage{multirow}
\usepackage{xspace} 

\makeatletter
\DeclareRobustCommand\onedot{\futurelet\@let@token\@onedot}
\def\@onedot{\ifx\@let@token.\else.\null\fi\xspace}

\makeatother

\begin{document}

\makeatletter
\def\@ACM@checkaffil{
    \if@ACM@instpresent\else
    \ClassWarningNoLine{\@classname}{No institution present for an affiliation}%
    \fi
    \if@ACM@citypresent\else
    \ClassWarningNoLine{\@classname}{No city present for an affiliation}%
    \fi
    \if@ACM@countrypresent\else
        \ClassWarningNoLine{\@classname}{No country present for an affiliation}%
    \fi
}
\makeatother

\title{TaleForge: Interactive Multimodal System for Personalized Story Creation}
\renewcommand{\shorttitle}{TaleForge: Personalized Story Creation} 


\author{Minh-Loi Nguyen}
\authornote{Equal contributions}
\orcid{0009-0003-2630-3325}
\affiliation{%
  \institution{\textsuperscript{\rm 1} University of Science, VNU-HCM, Ho Chi Minh City, Vietnam}
}
\affiliation{
  \institution{\textsuperscript{\rm 2} Vietnam National University, Ho Chi Minh City, Vietnam}
}

\author{Quang-Khai Le}
\authornotemark[1]
\orcid{0009-0004-3733-6496}
\affiliation{%
  \institution{\textsuperscript{\rm 1} University of Science, VNU-HCM, Ho Chi Minh City, Vietnam}
}
\affiliation{
  \institution{\textsuperscript{\rm 2} Vietnam National University, Ho Chi Minh City, Vietnam}
}

\author{Tam V. Nguyen}
\orcid{0000-0003-0236-7992}
\affiliation{%
  \institution{\textsuperscript{\rm 3} Department of Computer Science \\University of Dayton\\ Ohio, United States}
}

\author{Minh-Triet Tran}
\orcid{0000-0003-3046-3041}
\affiliation{%
  \institution{\textsuperscript{\rm 1} University of Science, VNU-HCM, Ho Chi Minh City, Vietnam}
}
\affiliation{
  \institution{\textsuperscript{\rm 2} Vietnam National University, Ho Chi Minh City, Vietnam}
}

\author{Trung-Nghia Le}
\authornote{Corresponding author. Email address: ltnghia@fit.hcmus.edu.vn}
\orcid{0000-0002-7363-2610}
\affiliation{%
  \institution{\textsuperscript{\rm 1} University of Science, VNU-HCM, Ho Chi Minh City, Vietnam}
}
\affiliation{
  \institution{\textsuperscript{\rm 2} Vietnam National University, Ho Chi Minh City, Vietnam}
}

\renewcommand{\shortauthors}{M.-L. Nguyen et al.}

\begin{abstract}
    Storytelling is a deeply personal and creative process, yet existing methods often treat users as passive consumers, offering generic plots with limited personalization. This undermines engagement and immersion, especially where individual style or appearance is crucial. We introduce TaleForge, a personalized story-generation system that integrates large language models (LLMs) and text-to-image diffusion to embed users’ facial images within both narratives and illustrations. TaleForge features three interconnected modules: Story Generation, where LLMs create narratives and character descriptions from user prompts; Personalized Image Generation, merging users’ faces and outfit choices into character illustrations; and Background Generation, creating scene backdrops that incorporate personalized characters. A user study demonstrated heightened engagement and ownership when individuals appeared as protagonists. Participants praised the system’s real-time previews and intuitive controls, though they requested finer narrative editing tools. TaleForge advances multimodal storytelling by aligning personalized text and imagery to create immersive, user-centric experiences.
\end{abstract}



\begin{teaserfigure}
  \centering
  \includegraphics[width=\textwidth]{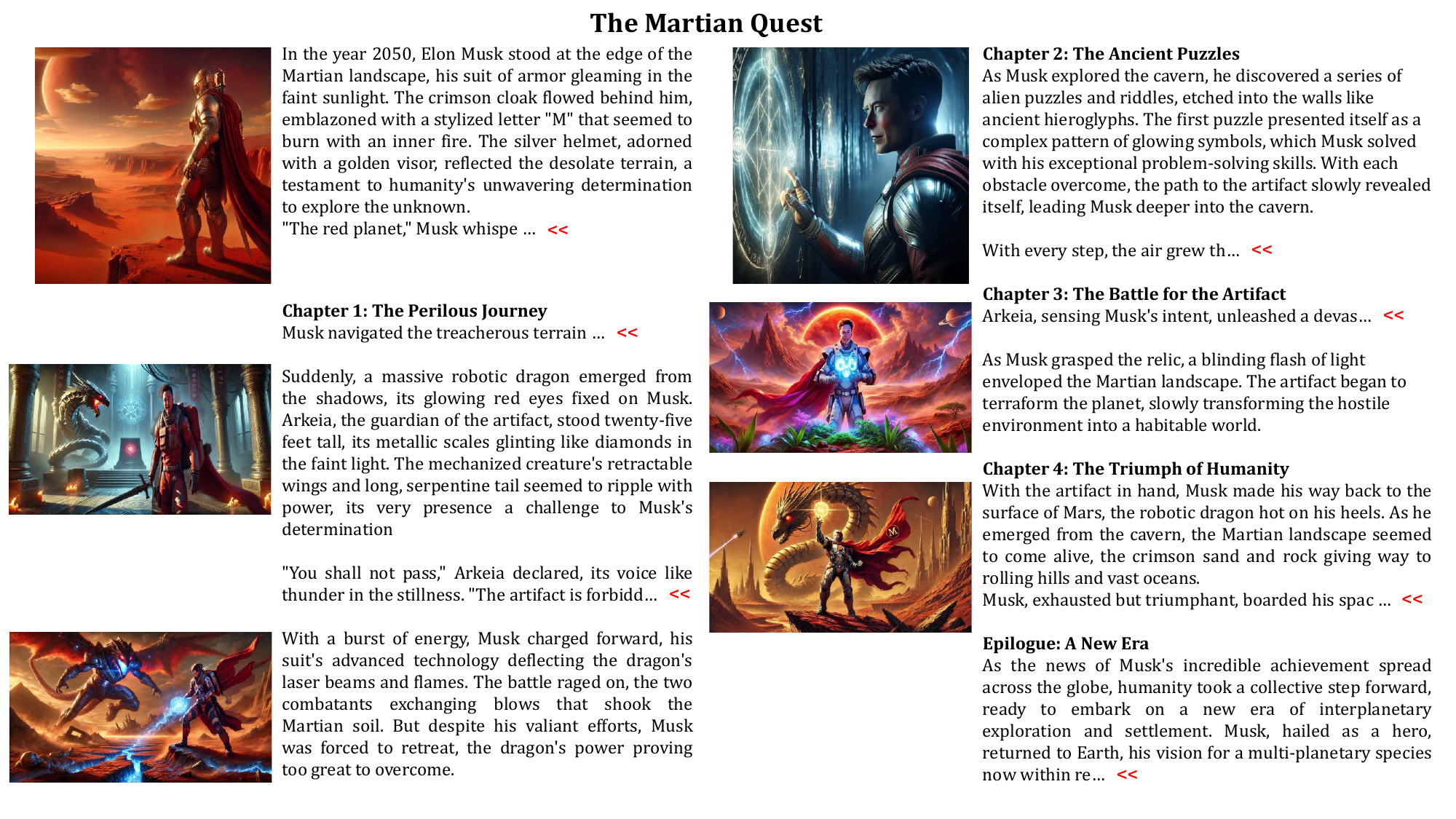} 
  \caption{Example storytelling generated by TaleForge. In this scenario, Elon Musk's face is uploaded and dressed in ‘Mars Style’ and ‘Knight Cloth’ armor, and he embarks on an epic quest. TaleForge creates personalized, high-quality images in which Elon Musk, as a knight, faces a robotic dragon in a Martian cavern while aligning with the story theme. The system seamlessly integrates the character and background, bringing to life his battle to retrieve a powerful artifact capable of terraforming Mars.}
  \Description{A series of four images depicting a story. The first image shows a portrait of Elon Musk. The second shows him in 'Mars Style' armor. The third shows him in 'Knight Cloth' armor. The final image shows Musk as a knight fighting a robotic dragon on Mars, illustrating the personalized story concept.}
  \label{fig:teaser}
\end{teaserfigure}

\maketitle

\section{Introduction}
\label{sec:introduction}

Storytelling is central to human communication, enabling the transmission of experiences, emotions, and cultural values. Recent advances in generative artificial intelligence (AI), particularly large language models (LLMs) and text-to-image diffusion models, have opened new opportunities for creating immersive, personalized narratives across domains such as entertainment \cite{wang2024weaver}, education \cite{laak2024ai}, and therapy. These technologies promise to actively engage users in story creation, fostering deeper personalization and connection.

Despite these advances, most existing systems offer generic, user-agnostic narratives, limiting users’ sense of immersion and ownership. While multimodal systems \cite{Kumari2023MultiConcept} integrate text and visuals, few incorporate personalized visual identities such as facial imagery, constraining narrative immersion. Although StoryDiffusion \cite{Zhou2024storydiffusion} enhances visual coherence and DreamBooth \cite{Ruiz2022DreamBoothFT} enables image personalization, comprehensive integration of user identity across both narrative and visuals remains underexplored.

To better understand user needs, we conducted a formative study comprising eight semi-structured interviews and an online survey with 20 participants, including teachers, hobbyist writers, and therapists. Participants overwhelmingly preferred simple, prompt-driven interfaces over complex design tools. These findings directly informed the design of TaleForge, emphasizing ease of use, real-time feedback, and intuitive personalization controls.

In this paper, we present TaleForge, an interactive storytelling platform that empowers users to personalize both the narrative and visual components of their stories. TaleForge allows users to upload personal images, select story settings and genres, and visualize themselves as protagonists through a three-stage generative pipeline. Story Generation uses LLMs to produce narrative outlines and detailed character descriptions from user prompts and genre preferences. Personalized Image Generation employs a multi-ControlNet refinement pipeline to accurately merge user face embeddings, clothing style, and pose into high-fidelity character renders. Background Generation synthesizes scene backdrops and supporting interactive bounding-box placement for coherent integration of personalized characters and environments.

A user study with 12 participants showed that representing users as protagonists enhanced engagement and ownership. Participants highlighted TaleForge’s real-time previews and intuitive controls while expressing interest in more precise narrative editing. The study confirmed strong usability and broad creative support across domains like game design, scripting, and content creation, with additional potential for social media storytelling in marketing and branding. Our contributions are threefold:
\begin{itemize}[nosep]
  \item We propose TaleForge, a novel personalized multimodal storytelling platform that combines narrative and visual personalization using advanced generative AI techniques to deliver immersive, user-centric storytelling experiences.
  \item We introduce a novel three-stage generative pipeline that systematically manages narrative creation, personalized character visualization, and coherent background composition.
  \item User study demonstrates system usability and effectiveness, offering insights for enhancing interactive personalization and narrative-visual coherence.
\end{itemize}

\section{Related Work}
Recent advances in large language models (LLMs) and generative AI have opened new possibilities for narrative generation. Models such as GPT \cite{brown2020language} have demonstrated the ability to generate coherent, context-aware narratives, contributing to applications in education, entertainment, and creative writing \cite{wang2024weaver, laak2024ai}. At the same time, developments in generative modeling, spanning GANs \cite{goodfellow2014generative}, VAEs \cite{kingma2013auto}, and diffusion models \cite{ho2020denoising}, have enabled high-quality image synthesis with increasingly fine-grained control. Tools such as Stable Diffusion \cite{Rombach_2022_CVPR} and DreamBooth \cite{Ruiz2022DreamBoothFT} offer user-specific visual customization via fine-tuning, enabling the creation of personalized avatars and scenes. However, these advancements often operate in isolation: language models produce text without visual grounding, and generative image models produce visuals without narrative context, limiting their potential for integrated, user-centered storytelling.

To bridge text and visuals, several recent efforts have explored interactive and personalized storytelling pipelines. Co-creating Visual Stories \cite{antony2023id8} and PatchView \cite{chung2024patchview} facilitate collaborative story creation, enabling users to co-author narratives with generative models. However, these systems focus primarily on story structure and textual collaboration, offering limited support for personalized or identity-aware visuals. CharacterMeet \cite{qin2024charactermeet} personalizes character interactions by enabling AI-driven dialogues tailored to user input, though it remains text-focused and does not incorporate dynamic visual representation. FairylandAI \cite{makridis2024fairylandai} introduces a template-driven framework for story customization yet lacks mechanisms for integrating personalized visual content. Other works have explored visual storytelling from a cultural perspective, such as Ferracani et al. \cite{ferracani2024personalized}, who emphasize the importance of cultural identity in visual curation but do not employ real-time generative systems. Meanwhile, approaches like StoryDiffusion \cite{Zhou2024storydiffusion} and StoryMaker \cite{Zhou2024StoryMakerTH} improve multimodal coherence across text and image modalities, and Kumari et al. \cite{Kumari2023MultiConcept} enable multi-concept visual synthesis, though such models typically lack deep personalization and do not reflect the unique characteristics of individual users.

In contrast to prior approaches, our TaleForge introduces a unified framework that integrates user identity into both narrative generation and visual synthesis. Its three-stage pipeline combines identity-aware prompting, personalized visual grounding, and multimodal alignment to produce coherent and customized storytelling experiences. This design supports coherent, personalized storytelling experiences that adapt to user traits, broadening the creative potential of generative AI systems.

\section{Proposed Method}
\label{sec:proposed_system}

\subsection{Overview}
\begin{figure}[t!]
    \centering
    \includegraphics[width=\linewidth]{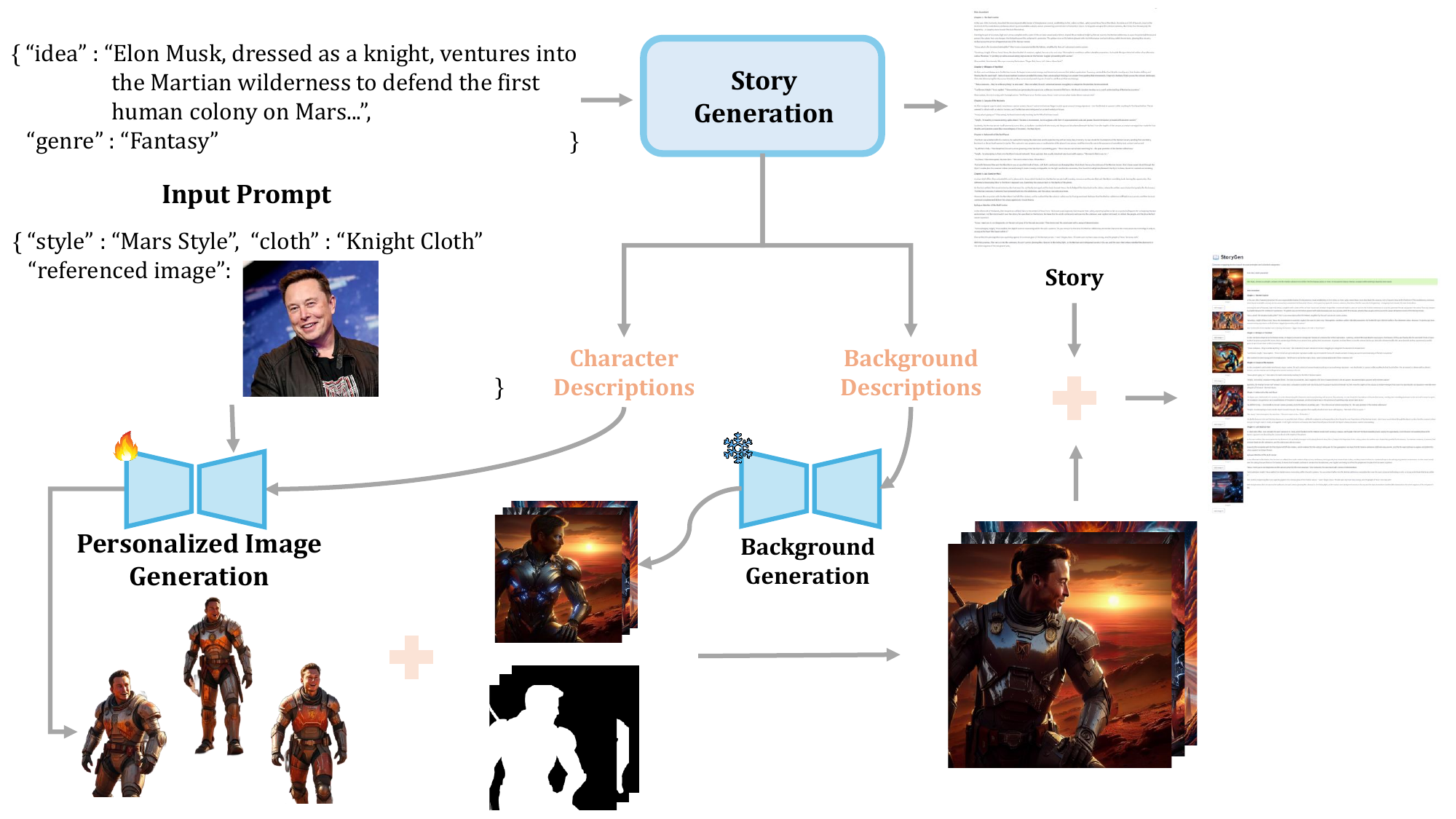}
    \caption{Overview of the proposed TaleForge framework, consisting of three main stages: (a) Story Generation, which creates a story based on the user's ideas and genre, while also generating additional details like Character Descriptions and Background Descriptions; (b) Personalized Image Generation, which creates images based on the user’s face and clothing, combined with the Character Descriptions from the Story Generation module; and (c) Background Generation, which creates images based on the Background Descriptions provided by the Story Generation module.}
    \Description{A pipeline diagram illustrating TaleForge. Stage (a) Story Generation takes user idea and genre as input, outputting story, character description, and background description. Stage (b) Personalized Image Generation takes user face, clothing, and character description, outputting a character image. Stage (c) Background Generation takes background description and produces a background image. These components combine to form the final story with images.}
    \label{Overview_method}
\end{figure}

Figure~\ref{Overview_method} shows an overview of TaleForge’s three-stage pipeline. The Story Generation module uses user inputs (story idea and genre) to generate a narrative along with metadata such as character and background descriptions. The Personalized Image Generation module combines user-provided facial images and clothing preferences with character sketches to create high-fidelity character visuals that fit with the generated story. The Background Generation module then synthesizes scenes based on the background description and integrates the personalized characters using bounding box placement. Finally, the system combines the generated images with the story paragraphs using key paragraphs provided by the Story Generation module.

\subsection{Story Generation}
TaleForge’s Story Generation module, powered by Llama3 \cite{grattafiori2024llama3herdmodels}, simultaneously produces a story outline and character sketches from user inputs. Generating these elements together ensures narrative coherence and prevents missing important characters or objects. The story outline and the character sketches are further processed by LLMs to create detailed character profiles, including physical features, attire, and poses, supporting accurate image generation. Metadata, including key story paragraphs and detailed scene descriptions (e.g., characters and background descriptions), further guides illustration generation.

\begin{figure}[t!]
    \centering
    \begin{subfigure}[b]{\linewidth}
        \includegraphics[width=\textwidth, trim=0pt 210pt 0pt 0pt, clip]{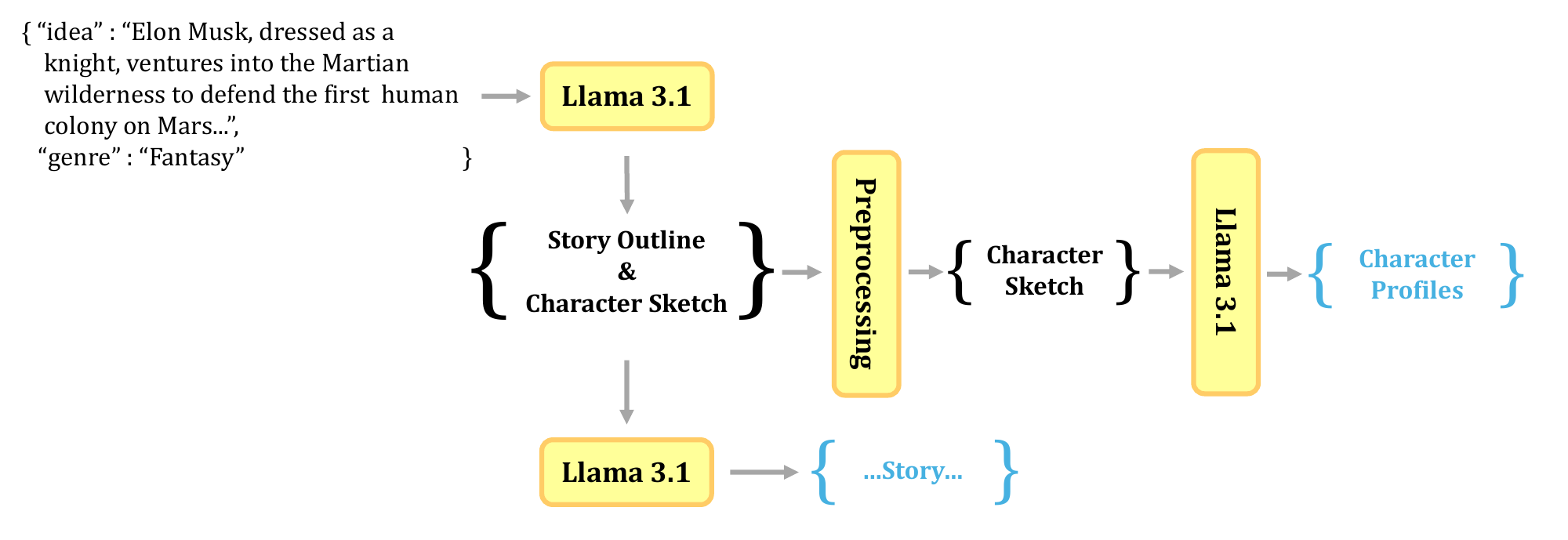}
        \caption{Raw Story}
        \label{fig:rawstory}
    \end{subfigure}
    \begin{subfigure}[b]{\linewidth}
        \includegraphics[width=\textwidth,  trim=0pt 130pt 0pt 0pt, clip]{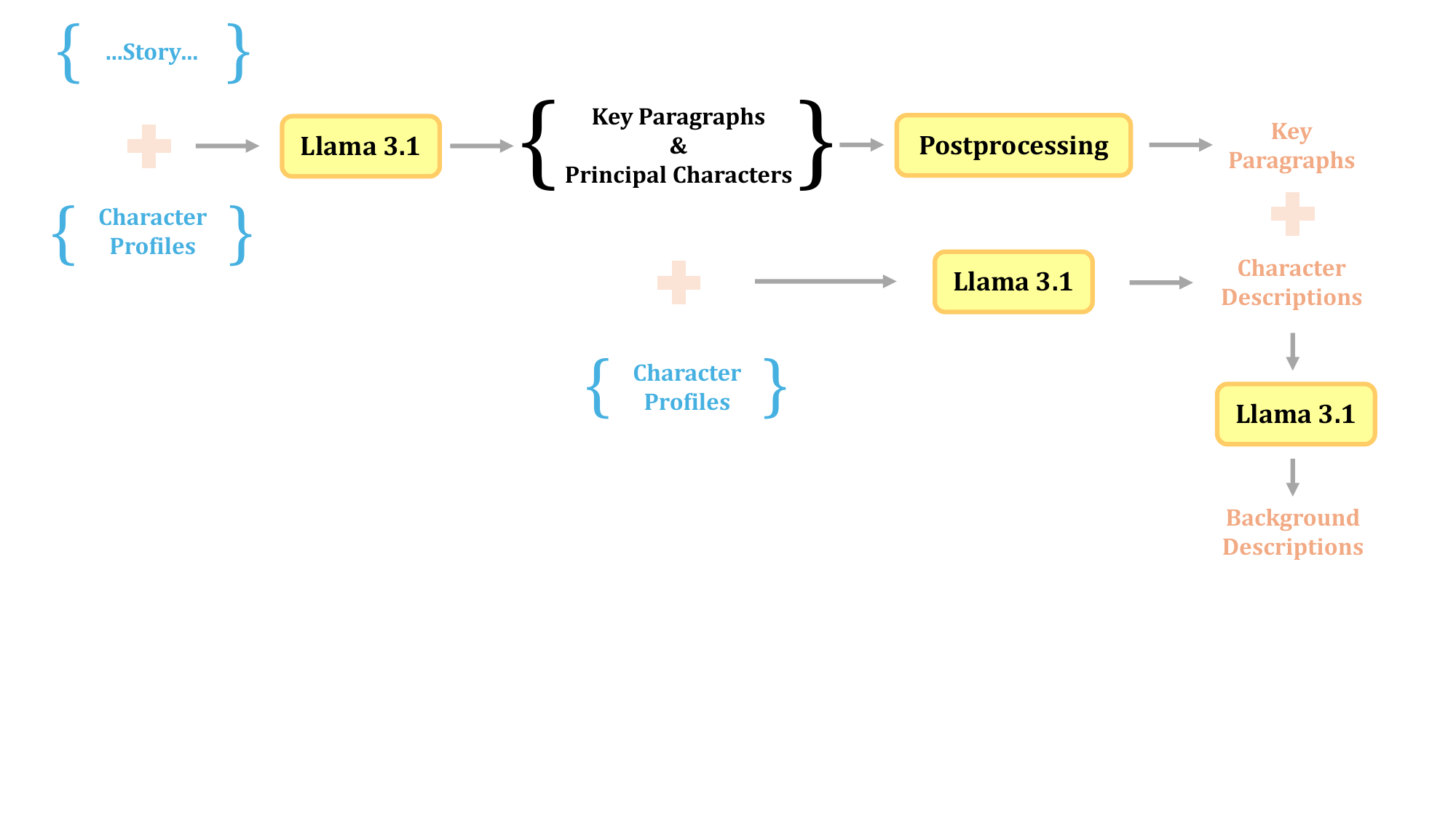}
        \caption{Metadata with the story}
        \label{fig:metadata}
    \end{subfigure}
    \caption{Pipeline of the Story Generation module. (a) User provides a story idea and genre, which Llama3 processes into a raw story. (b) LLMs then extract character and background descriptions as metadata from the raw story.}
    \Description{Two-part diagram of Story Generation. Part (a) shows inputs (User Idea, Genre) feeding into Llama3, which outputs a 'Raw Story'. Part (b) shows the 'Raw Story' feeding into LLMs, which output 'Character Description', 'Background Description', and 'Key Paragraphs'.}
    \label{StoryGeneration_method}
\end{figure}

\subsection{Personalized Image Generation}
The Personalized Image Generation module integrates user inputs, such as facial references, clothing styles, and poses, through a two-stage pipeline leveraging StoryMaker \cite{Zhou2024StoryMakerTH} and InstantID \cite{Wang2024InstantIDZI}. User inputs are preprocessed using pose estimation and semantic segmentation techniques to ensure accurate user-defined attribute integration. Using these inputs, StoryMaker \cite{Zhou2024StoryMakerTH} creates initial images by blending user facial features with selected garments. InstantID \cite{Wang2024InstantIDZI} then refines these outputs, applying stylistic filters and precise pose adjustments through ControlNet and Multi-ControlNet. This pipeline ensures consistency in style and fidelity while preserving user identity.

\subsection{Background Generation}
The Background Generation module creates scenes from story-derived background descriptions using DALL-E 3. User-drawn bounding boxes and segmentation with SAM \cite{kirillov2023segany} partition the background to insert characters naturally. DreamBooth \cite{Ruiz2022DreamBoothFT} is used to fine-tune Stable Diffusion \cite{Rombach_2022_CVPR} for character style learning. Finally, the newly fine-tuned weights are used in the Swap Anything model \cite{gu2024swapanything} to insert the characters seamlessly into the background scene.

\begin{figure}[t!]
    \centering
    \begin{subfigure}[b]{\linewidth}
        \includegraphics[width=\textwidth, trim=0pt 50pt 0pt 0pt, clip]{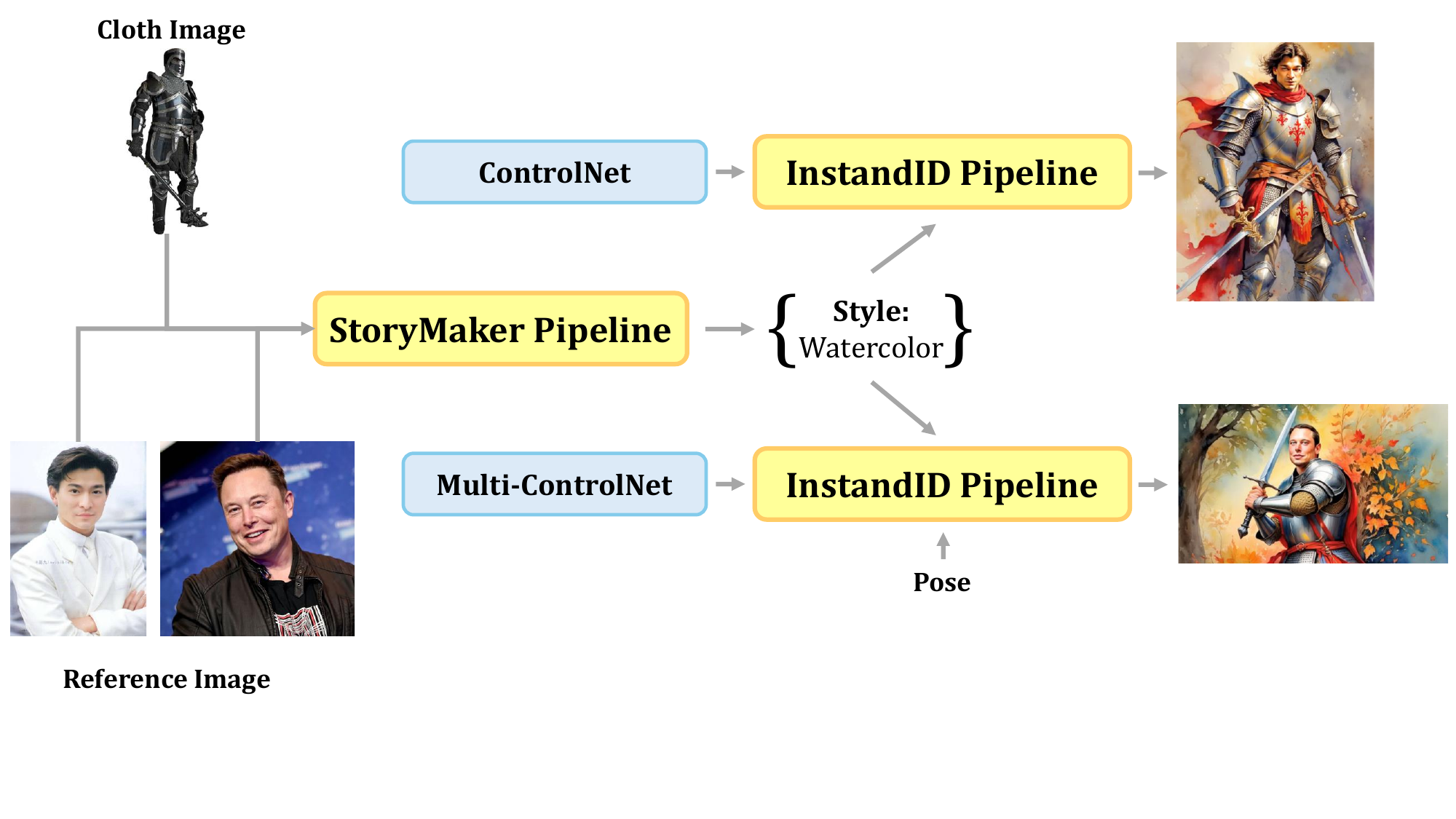}
        \caption{Overview of image generation workflow with two main capabilities}
        \label{fig:Character Generation}
    \end{subfigure}
    \begin{subfigure}[b]{\linewidth}
        \includegraphics[width=\textwidth,  trim=0pt 130pt 0pt 0pt, clip]{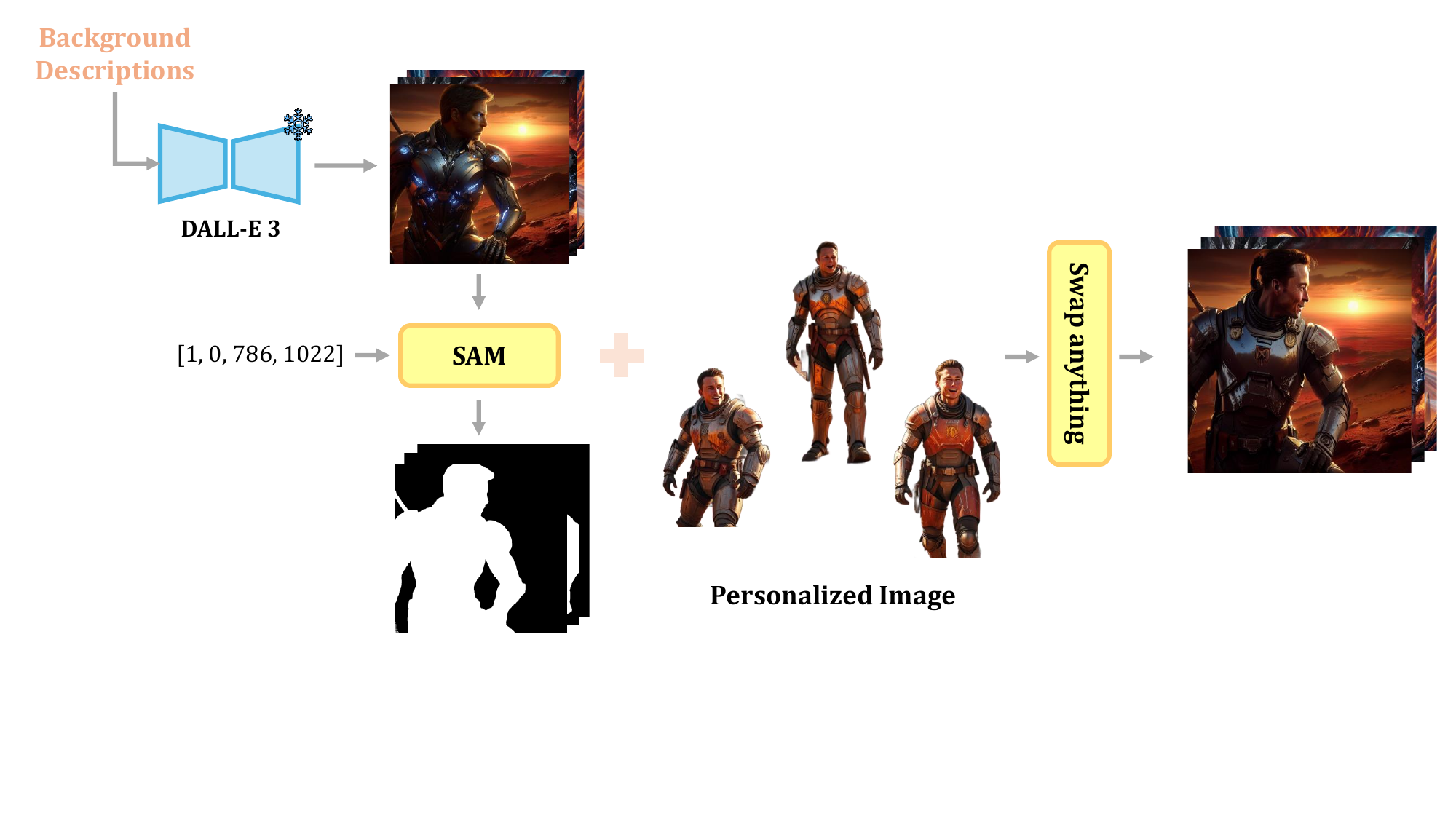}
        \caption{Pipeline of Background Generation module.}
        \label{backgroundGen_method}
    \end{subfigure}
    \caption{Overview of visual synthesis process. (a) The Personalized Image Generation module transforms user inputs (face, clothing, pose, style) via StoryMaker and InstantID into a character image. (b) The Background Generation module uses background descriptions with DALL-E 3, SAM, DreamBooth, and Swap Anything to create scenes and integrate characters.}
    \Description{Two-part diagram for visual synthesis. (a) Personalized Image Generation: inputs (Face, Clothing, Pose, Style) go to StoryMaker, then InstantID (with ControlNet/Multi-ControlNet) to produce a Character Image. (b) Background Generation: input 'Background Description' goes to DALL-E 3, then SAM with user bounding box, then DreamBooth fine-tuning Stable Diffusion, then Swap Anything to combine with the character image into a final scene.}
    \label{ImageGeneration_method}
\end{figure}

\section{User study}
We conduct a user study to gather feedback on our system. We focus on evaluating how well the system provides a smooth and enjoyable experience, improving both the story and the visual elements.

\subsection{Setup}
We recruited 12 participants (aged 18–35, with varying digital and AI experience) to evaluate TaleForge. Participants first interacted with a baseline system featuring default characters and stories, then used TaleForge’s personalized features. Participants rated the generated stories and images from 1 (very bad) to 5 (very good) across seven criteria:
\begin{itemize}[nosep] 
    \item \textbf{Face Similarity}: assessing how closely the generated character’s face matches the uploaded face.
    \item \textbf{Garment Consistency}: evaluating the alignment and consistency of garments across images.
    \item \textbf{Character Integrity}: focusing on the completeness and correctness of the character's body.
    \item \textbf{Story-Concept Alignment}: measuring how well the story matches the input idea and genre.
    \item \textbf{Story Engagement}: assessing the appeal and ability of the story to captivate participants.
    \item \textbf{Image-Story Relevance}: evaluating the correspondence between images and story content.
    \item \textbf {Visual Naturalness}: focusing on the seamless integration of characters into background images without distortion or unrealistic alterations.
\end{itemize}

\subsection{Tasks}
Participants completed four tasks, each emphasizing different personalization elements: (1) single-character stories, (2) single-character with an object, (3) dual-character stories, and (4) single-character with two objects. Tasks varied genres and styles to assess TaleForge’s flexibility.
\begin{itemize}[nosep]
\item \textbf {Task 1: Single Person.} Participants upload an image of Elon Musk and select "Mars Style" and "Knight Cloth." The story shows Elon Musk as a knight exploring Mars and protecting the first human colony while facing strange Martian creatures with a laser sword.
\item \textbf{Task 2: Single Person and One Cameo Object.} Participants upload an image of Andy Lau Tak-wah, select "Watercolor Style" and "Astronaut Cloth." The story follows Andy, in an astronaut suit, discovering a living alien soccer ball and working together to decode an ancient alien language.
\item \textbf{Task 3: Two Persons.} Participants upload images of Elon Musk and Andy Lau Tak-wah, choose the "Watercolor Style" and "Knight Cloth" for both. The story features Elon Musk and Andy Lau Tak-wah teaming up to save Earth from a massive asteroid. Elon devises a high-tech solution, while Andy Lau Tak-wah uses his knightly skills to navigate an ancient temple that holds the key to destroying the asteroid. Their contrasting styles lead to humorous and tense moments.
\item \textbf{Task 4: Single Person + 2 Cameo Objects.} Participants upload an image of Elon Musk and select "Mars Style" and "Knight Cloth." The story depicts Elon Musk, dressed as a knight, battling a robotic dragon guarding a powerful Martian artifact. The artifact has the ability to terraform Mars, but it is hidden in a perilous alien cavern. Elon must outsmart the dragon and solve alien puzzles to retrieve the artifact.
\end{itemize}

\begin{table}[t!]
  \caption{Mean participant ratings (1–5) across four multimodal story generation tasks. Higher scores indicate more favorable perceptions.}
  \label{study_metrics}
  \setlength{\tabcolsep}{4pt} 
  \begin{tabularx}{\linewidth}{ 
      p{0.35\linewidth}                    
      >{\centering\arraybackslash}X       
      >{\centering\arraybackslash}X       
      >{\centering\arraybackslash}X       
      >{\centering\arraybackslash}X       
      }
    \toprule
    \textbf{Aspect}
      & \textbf{Task 1}
      & \textbf{Task 2}
      & \textbf{Task 3}
      & \textbf{Task 4} \\
    \midrule
    Story‑Concept Alignment  & 4.25 & 4.25 & 3.58 & 4.25 \\ 
    \midrule
    Story Engagement          & 4.25 & 4.25 & 3.75 & 4.42 \\ 
    \midrule
    Face Similarity           & 3.58 & 3.50 & 3.33 & 3.67 \\ 
    \midrule
    Garment Consistency       & 4.00 & 4.33 & 3.42 & 4.12 \\ 
    \midrule
    Character Integrity       & 3.83 & 3.08 & 2.92 & 3.58 \\ 
    \midrule
    Image‑Story Relevance     & 4.00 & 3.83 & 3.33 & 3.67 \\ 
    \midrule
    Visual Naturalness        & 3.92 & 3.75 & 2.83 & 3.41 \\ 
    \bottomrule
  \end{tabularx}
\end{table}

\subsection{Apparatus and Procedure}
The user study took place in a controlled laboratory environment. Each participant used a laptop pre-configured with the story creation system and performed tasks under supervision. To reduce bias from previous system experience, we divide participants into two groups. The first group starts with Task 1, and the second group starts with Task 2. This helped balance out any learning effects that could skew the results as participants became more familiar with the functionalities of the system over time. 
Each task was designed to be completed sequentially, allowing users to interact with the system as needed. After completing the tasks, participants were asked to provide feedback to gather qualitative insights into their experience. The average duration of each study session, including the tasks and post-task interview, was approximately 60 minutes.

\subsection{Insights}
Table~\ref{study_metrics} summarizes participant ratings across four progressively complex story generation tasks involving personalized image and narrative synthesis. TaleForge consistently achieved high scores in Story–Concept Alignment (4.25) across most tasks, indicating reliable translation of abstract user intent into coherent narrative structure (e.g., "Elon Musk as a knight on Mars" or "Andy Lau with an alien soccer ball"). Engagement scores were similarly strong, with generated stories capturing the intended genre and stylistic framing, including thematic consistency with user-selected visual motifs (e.g., Mars Style, Watercolor).

Personalization was well-received. In single-character settings (Tasks 1 and 2), Face Similarity and Character Integrity scored positively, with participants recognizing identity preservation even under stylized distortions (e.g., knight armor or watercolor rendering). Visual consistency deteriorated in Task 3, where multi-character alignment resulted in reduced Character Integrity (2.92) and Visual Naturalness (2.83), indicating that TaleForge currently faces challenges in maintaining inter-character consistency during complex interactions.

Garment fidelity remained relatively robust across conditions, particularly in Task 2 (4.33), where clothing attributes were preserved across scenes. However, action-heavy prompts (e.g., dragon battles in Task 4) introduced pose-dependent artifacts (e.g., limb warping, edge smearing), suggesting instability in high-motion synthesis.

Generated backgrounds were typically relevant to the story context (e.g., Martian landscapes, alien temples), supporting immersive scene composition. Still, visual-naturalness gaps were noted in dynamic or multi-entity scenes due to occasional compositional inconsistencies.

Interview feedback also highlighted the system’s intuitive interface and fast visual feedback. Interview responses emphasized the creative empowerment afforded by protagonist embodiment, aligning with prior findings in interactive storytelling. Use cases spanned game narrative prototyping, software demonstration generation, and marketing asset creation. In particular, participants noted TaleForge’s efficiency in producing character-driven social media content, suggesting applicability to automated brand storytelling pipelines.

These findings validate TaleForge’s feasibility for real-time, multimodal content creation while underscoring current limitations in scalability to multi-character coherence, dynamic realism, and artifact suppression.

\subsection{Discussions}

\subsubsection{Limitations}
TaleForge shows potential in generating personalized stories and images but has some limitations. First, the personalized image generation module occasionally fails to meet user expectations, with characters sometimes showing facial or body distortions that affect the integration of character images into backgrounds, creating a lack of coherence. Second, the Background Generation module struggles with complex background descriptions from the Story Generation module. Models like DALL-E 3, BingAI, and Midjourney may fail to generate backgrounds that fully capture the detailed context, causing a disconnect between the story and its visual representation. These issues underscore the need for further refinement in both image generation and the integration of story and visuals to improve overall quality and consistency. 

\subsubsection{Directions for Future Work}
Limitations found in the use study can be addressed by incorporating granular editing tools that allow selective paragraph or scene revisions without disrupting overall coherence \cite{wang2024weaver, laak2024ai}. We also aim to explore advanced depth estimation and image-segmentation methods \cite{kirillov2023segany} for integrating personalized characters into complex backgrounds and investigate specialized art styles (e.g., Impressionist, Baroque) through stroke-based generation models \cite{painttransformer}.

\section{Conclusion}
\label{sec:conclusion}
We presented TaleForge, a personalized story-generation system that integrates LLMs and text-to-image diffusion to create immersive, user-centric narratives. Through a three-stage pipeline, TaleForge enables users to visually depict themselves as protagonists. The user study highlighted our system’s ease of use, effective personalization, and potential for broader creative applications. Future work will address visual consistency challenges and explore finer control over narrative customization to further enhance user experience.


\begin{acks}
This research is funded by Vietnam National Foundation for Science and Technology Development (NAFOSTED) under Grant Number 102.05-2023.31.
\end{acks}

\bibliographystyle{ACM-Reference-Format}
\bibliography{ref} 

\end{document}